\documentclass[letterpaper]{article} 
\usepackage{aaai2026}  
\usepackage{times}  
\usepackage{helvet}  
\usepackage{courier}  
\usepackage[hyphens]{url}  
\usepackage{graphicx} 
\usepackage{natbib}  
\usepackage{caption} 
\usepackage{algorithm}
\usepackage{algorithmic}
\usepackage{amsmath}
\usepackage{pifont}
\usepackage{booktabs}
\usepackage{newfloat}
\usepackage{listings}
\DeclareCaptionStyle{ruled}{labelfont=normalfont,labelsep=colon,strut=off} 
\floatstyle{ruled}
\newfloat{listing}{tb}{lst}{}
\floatname{listing}{Listing}
\title{RoboRetriever: Single-Camera Robot Object Retrieval via Active and Interactive Perception with Dynamic Scene Graph}

\author {
    Hecheng Wang\textsuperscript{\rm 1},
    Jiankun Ren\textsuperscript{\rm 1},
    Jia Yu\textsuperscript{\rm 1},
    Lizhe Qi\textsuperscript{\rm 1},
    Yunquan Sun\textsuperscript{\rm 1}
}
\affiliations {
    \textsuperscript{\rm 1}College for Elite Engineer, Fudan University\\
}

\usepackage{bibentry}

\begin{document}

\maketitle

\begin{abstract}
Humans effortlessly retrieve objects in cluttered, partially observable environments by combining visual reasoning, active viewpoint adjustment, and physical interaction—with only a single pair of eyes. In contrast, most existing robotic systems rely on carefully positioned fixed or multi-camera setups with complete scene visibility, which limits adaptability and incurs high hardware costs. We present \textbf{RoboRetriever}, a novel framework for real-world object retrieval that operates using only a \textbf{single} wrist-mounted RGB-D camera and free-form natural language instructions. RoboRetriever grounds visual observations to build and update a \textbf{dynamic hierarchical scene graph} that encodes object semantics, geometry, and inter-object relations over time. 
The supervisor module reasons over this memory and task instruction to infer the target object and coordinate an integrated action module combining \textbf{active perception}, \textbf{interactive perception}, and \textbf{manipulation}. To enable task-aware scene-grounded active perception, we introduce a novel visual prompting scheme that leverages large reasoning vision-language models to determine 6-DoF camera poses aligned with the semantic task goal and geometry scene context. We evaluate RoboRetriever on diverse real-world object retrieval tasks, including scenarios with human intervention, demonstrating strong adaptability and robustness
in cluttered scenes with only one RGB-D camera.
\end{abstract}

\section{Introduction}
\begin{figure}[t]
\centering
\includegraphics[width=0.45\textwidth]{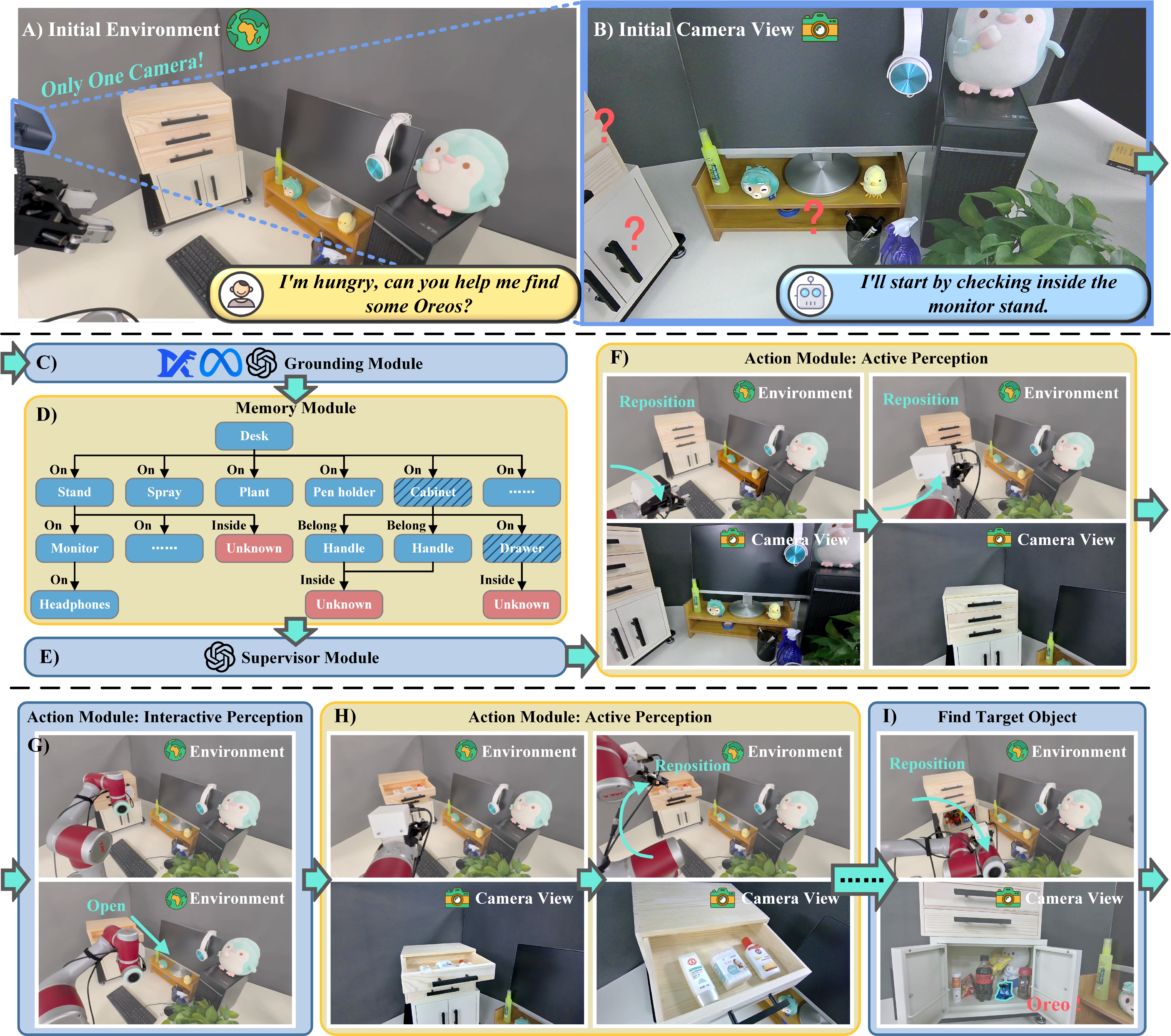} 
\caption{\textbf{RoboRetriever.} We propose a novel robotic framework for general-purpose object retrieval with only a single wrist-mounted RGB-D camera. Given a free-form user instruction, it first grounds observation and build a dynamic hierarchical scene graph capturing object semantics, geometries, and inter-object relations. 
Guided by its iteratively updated memory and task instruction, the system autonomously coordinates task-aware scene-grounded active perception, interactive perception and manipulation actions to explore and retrieves the target object
.}
\label{fig:intro}
\end{figure}

Imagine standing at your cluttered office desk, feeling a bit hungry and looking for a snack (e.g., Figure \ref{fig:intro} A). You recall that there are Oreos somewhere, but they’re not in your immediate view. As a human, you effortlessly hypothesize where it might be—perhaps inside a drawer or behind a stack of books—then move your head, lean over, or open a drawer to visually confirm. Despite having only a single pair of eyes, you seamlessly interleave head movements and physical interactions with the environment to locate your target. 
This seemingly simple behavior presents a significant challenge for embodied AI: How can we enable robots, equipped with only a single wrist-mounted RGB-D camera like the human eye, to autonomously infer where to look, how to interact, and what to manipulate—based solely on a free-form user instruction—in order to retrieve a target object in a complex, partially observable scene?


By contrast, most robotic systems sidestep this problem by assuming far more permissive perception conditions: either using a carefully placed camera that can observe all task-relevant content\cite{pushing2, opening4}, or deploying  multiple redundant cameras to cover the workspace\cite{multi1,multi2}. 
However, their adaptability is limited, as changes in task objectives or environment can significantly affect the optimal viewpoint. 
Recent research in interactive perception\cite{RoboExp,pushing2} has attempted to address the partial observability problem by enabling robots to acquire information through physical interaction—such as opening containers or moving occlusions. However, such methods typically operate under carefully designed, fixed-camera setups and lack the ability to autonomously adjust the robot's perspective to gather additional information. In parallel, active perception\cite{navi4, navi8} allow robots to control camera viewpoints, but are mostly developed for mobile robot navigation in 2D space, rather than for 6-DoF manipulation tasks where occlusion and spatial constraints are far more complex. For example, human may need to bend down and reorient their gaze to inspect the inside of a cabinet. Critically, existing active perception strategies in the manipulation domain are often hardcoded\cite{hard} or task-specific\cite{learn3,grasp2}, and lack the ability to interpret and follow open-ended language goals, limiting adaptability to diverse instructions and scenes.


To tackle this challenge, we propose a novel real-world robotic object retrieval framework, \textbf{RoboRetriever}, along with a novel visual prompting scheme for robot active perception, and a novel dynamic hierarchical scene graph representation. The framework comprises four modules: Grounding Module, Memory Module, Supervisor Module, and Action Module, illustrated in Figure \ref{fig:intro}. The Grounding Module first leverages GPT-4o, DINO-X\cite{DINO}, and Segment-Anything(SAM)\cite{SAM} to segment objects from RGB-D images, extracting detailed geometric information. 
The Memory Module maintains a dynamic hierarchical scene graph in which each object is represented as a node with semantic and geometric attributes, and inter-object relationships are encoded as edges. The memory is incrementally updated with new observations, enabling a progressively refined understanding of the environment. 
The Supervisor Module reasons over the current memory and task instruction to infer the target object and appropriate action. The Action Module includes three capabilities: active perception, interactive perception, and manipulation. For active perception, we propose a novel visual prompting scheme that leverages GPT-o3 in a two-stage procedure to determine the 6D camera pose based on the scene point cloud, target object, and camera parameters. In addition, we implement a set of action primitives to enable robots to physically interact with the environment.

We validate our method across a diverse categories of real-world object retrieval tasks with varying attributes and complexities, demonstrating the adaptability and robustness of our system. Our contributions are as follows:
(1) We propose a novel robotic framework that dynamically updates a graph-based memory and hypothesizes unseen areas, and coordinates active and interactive perception strategies for effective object retrieval with only a single RGB-D camera.
(2) We introduce a novel visual prompting scheme that bridges natural language task specifications with grounded 3D scene knowledge. This enables a reasoning vision-language model(VLM) to perform task-aware scene-grounded active perception.
(3) We design a hierarchical dynamic scene graph that continuously updates to capture semantic, geometric, and relational information of objects, enabling progressively  refined scene understanding.

\section{Related Work}
\noindent \textbf{Interactive Perception for Robotics.}
Interactive perception enables robots to better understand their environments by coupling perception with physical interaction. Previous works have utilized predefined action primitives, such as poking\cite{poking, poking2}, pushing\cite{pushing1, pushing2, pushing3}, and opening\cite{opening1, opening2, opening3}, to expose occluded objects or discover unknown structures. For example, poking-based approaches \cite{poking} reconstruct unseen 3D objects through  multi-view observations during interaction, bypassing the need for prior object models or category-specific training. Recent works, such as RoboEXP\cite{RoboExp}, CuriousBot\cite{curiousbot}, and Structure-from-Action (SfA)\cite{SfA}, extend this by constructing rich scene graphs or articulated CAD models via sequential interactions, revealing both spatial and action-conditioned relationships. Other works have also integrated uncertainty-aware action selection and hierarchical policies\cite{pushing3} to guide interaction more efficiently in cluttered scenes. However, these systems often rely on carefully positioned or multi-camera setups, operate in structured environments, or lack adaptability to unseen, language-driven tasks. 
In contrast, our method unifies active and interactive perception under a single-camera constraint and adapts across diverse tasks and environments without task-specific priors.
 
\noindent \textbf{Active Perception for Robotics.}
Active perception empowers robots to autonomously select informative viewpoints to reduce uncertainty and improve scene understanding. Prior work has primarily focused on mobile robots in navigation tasks\cite{navi1,navi2,navi3}. 
In manipulation contexts, active perception has been applied to specific tasks such as grasp refinement\cite{grasp1,grasp2,grasp3}, bin-picking\cite{bin1,bin2}, or object pose estimation\cite{pose1,pose2}, often through viewpoint selection heuristics\cite{huris}, information gain maximization\cite{gain1, gain2}, or policy learning\cite{learn1,learn2,learn3}. However, these approaches typically assume static scenes, rely on predefined object categories, and are tightly coupled with the downstream task, making them difficult to adapt to unseen tasks and scenes. More critically, they lack the ability to interpret high-level semantic goals expressed in natural language, limiting their adaptability and versatility. In contrast, our approach introduces a general-purpose active perception framework that seamlessly bridges natural language task objective with grounded 3D scene knowledge. By prompting a reasoning VLM, our system interpret free-form language instructions and conduct task-aware scene-grounded active perception in a plug-and-play way.


\section{Method}

\begin{figure*}[t]
\centering
\includegraphics[width=0.8\textwidth]{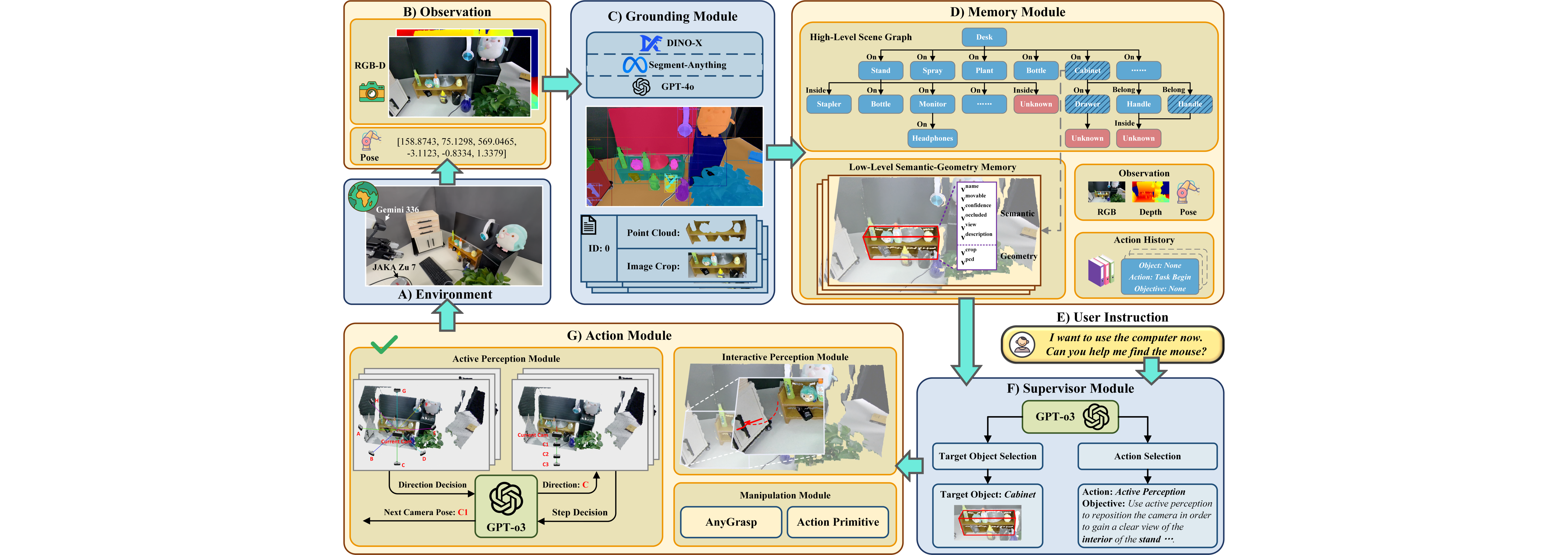} 
\caption{\textbf{Overview of our proposed system.} 
The system consists of four key modules
: grounding module, memory module, supervisor module, action module. The grounding module processes observations to extract geometric information. The memory module maintains a dynamic high-level scene graph, integrated with low-level semantic-geometry memory, current observations, and action history. Based on the memory and task instruction, the supervisor module infers the target object, next action and corresponding goal description. 
The action module supports three types of actions.
Active perception enable system to take active control of its perception. Interactive perception enable system to acquire task-relevant information by physically interacting with objects. 
Once sufficient information is acquired, the system take manipulation action to retrieve the object.
}
\label{overview}
\end{figure*}

\subsection{Problem Formulation}
In this work, we focus on addressing the object retrieval problem under limited perception conditions. In this setting, the robotic manipulation system is equipped with only one RGB-D camera mounted on wrist and a parallel-jaw gripper, operating in an unknown, complex environment that resembles real-world scenarios. The user issues a free-form language instruction $l \in \mathcal{L}$ that specifies the need for a target object $\omega^* \in \Omega$. The target object $\omega^*$ is not visible in the robot's initial observation $o_0$. 

We formalize the agent as $\Pi = \{\mathcal{G}, \mathcal{M}, \mathcal{D}, \mathcal{ACT}\}$, where $\mathcal{G}$ is the grounding module, $\mathcal{M}$ is the memory module, $\mathcal{D}$ is the supervisor module, $\mathcal{ACT}$ is the action module. As shown in Figure \ref{overview}, at time $t$, the grounding module processes the current observation $o_t$ to extract information $\mathcal{I}_t$, $\mathcal{I}_t = \mathcal{G}(\cdot|o_t)$. 
The memory module then updates the internal memory based on the previous memory, the current observation, and the extracted information, 
$\mathcal{M}_{t} = \mathcal{U}(\cdot|\mathcal{M}_{t-1}, o_t, \mathcal{I}_t)$. Given the user instruction $l$ and the updated memory $\mathcal{M}_t$, the supervisor module $\mathcal{D}$ infers the target object and high-level action $a^H_t$. It also generates an intermediate natural language description $g_t$ to explain action objective, $a^H_t, g_t = \mathcal{D}(\cdot|l, \mathcal{M}_t)$. Finally, the action module generates the low-level action based on the high-level action and memory $a^L_t = \mathcal{ACT}(\cdot|a^H_t, \mathcal{M}_t)$, which is then executed by the robot. 

\subsection{Grounding module}
After receiving the current observation $o_t$, consisting of the RGB image, depth image, and robot pose, the grounding module extracts useful information to support subsequent memory updates, as illustrated in Figure \ref{overview} C. Specifically, we first project the RGB-D image into a 3D point cloud and transform it into the base frame using hand-eye calibration and robot pose. Then we leverage GPT-4o to infer object classes and use DINO-X\cite{DINO} and SAM\cite{SAM} to obtain object bounding boxes and segmentation masks. These are used to obtain object segmented point cloud $\omega^{pcd}_{t, i}$ and cropped image $\omega^{crop}_{t, i}$.
Finally, the output of grounding module is constructed as 
\begin{equation}
\mathcal{I}_t = \{I_{t,0}, I_{t,1}, ..., I_{t,N}\}
\end{equation}
\begin{equation}
I_{t,i} = \{ \omega^{crop}_{t, i}, \omega^{pcd}_{t, i}\}
\end{equation}

\subsection{Memory module}
The memory module consolidates scene information and continuously updates to maintain consistency with the evolving environment. It comprises three components: a dynamic hierarchical scene graph, current observation, and system action history, as illustrated in Figure \ref{overview} D. We define the dynamic hierarchical scene graph as $G = (V, E)$, where $V=\{v_0, v_1, ..., v_N\}$ represents object nodes, and $E = \{e_0, e_1, ..., e_N\}$ represents relation edges. Each node either links to a known low-level semantic-geometric object entry, or represents an \textit{Unknown} node hypothesizing potential objects in unexplored areas. 
Each edge denotes a directed spatial or semantic relationship between two objects. The action history records prior agent decisions, including target objects, actions, and associated goal descriptions.


To handle dynamic environments, the memory module is updated with each new observation.
Specifically, for each detected object $I_{t,i}$ at time $t$, we match it to existing nodes $v_j$ in the scene graph using GPT-4o. The model is prompted with the current object cropped image $\omega^{crop}_{t,i}$ and historical cropped images $v^{crop}_{0...t-1,j}$ to assess instance correspondence.

If a match is found, the current cropped image $\omega^{crop}_{t,i}$ is appended to the historical sets of node $v_j$. We then apply the ZeroMatch\cite{zero} algorithm to register the current point cloud $\omega^{pcd}_{t,i}$ with the accumulated point cloud $v^{pcd}_{t-1,j}$, producing a refined point cloud $v^{pcd}_{t,j}$. Next, we leverage GPT-o3 to update node's semantic attributes based on its image history, including:
\begin{itemize}
    \item $v^{name}_{t,j}$: a fine-grained semantic label.
    \item $v^{movable}_{t,j}$: a binary indicator of whether the object is movable or static.
    \item $v^{conf}_{t,j}$: a confidence score reflecting the reliability of the semantic label.
    \item $v^{occl}_{t,j}$: a binary indicator of whether the object has remained occluded across all historical observations.
    \item $v^{view}_{t,j}$: a binary indicator of whether the object has remained partially observed across all frames due to camera viewpoint limitations.
    \item $v^{desc}_{t,j}$: a free-form sentence describing the object.
\end{itemize}
If $I_{t,i}$ does not match any existing node, a new node is created and initialized using the same procedure. The memory entry for node $v_j$ at time $t$ is defined as:
\begin{equation}
\begin{split}
v_{t,j} = \{ & v^{crop}_{t,j}, v^{pcd}_{t,j}, v^{name}_{t,j}, v^{movable}_{t,j}, \\
                   & v^{conf}_{t,j}, v^{occl}_{t,j}, v^{view}_{t,j}, v^{desc}_{t,j} \}
\end{split}
\end{equation}

We define a set of rules that integrate geometric, semantic, and action-based information to establish relationships between objects with the help of VLM. 
Geometric attributes (e.g., bounding box) support spatial relations such as on or under. Semantic labels help infer part-whole relations (e.g., a handle belongs to a cabinet). Action history also informs relations—e.g., lifting one object to reveal another implies the latter was under the former.
Our graph encodes five types of relations: behind, belong, inside, on, and under.


Relations such as behind, inside, and under often indicate occluded or unexplored regions (e.g., a closed drawer associated with “inside”). For such cases, we assess whether a corresponding \textit{Unknown} node should be introduced. We prompt the GPT-4o to reason about whether such a node should be added, based on relational context and scene understanding. If such a hypothesis is supported, we instantiate a \textit{Unknown} node and attach it accordingly in the graph.



\subsection{Supervisor module}
As illustrated in Figure \ref{overview} F, the supervisor module is responsible for determining which object the system should act upon and what type of action should be executed, based on the task instruction. 
It ensures local goal consistency across the action history, enabling coherent and purposeful behavior.
We employ GPT-o3, a state-of-the-art reasoning VLM, as the core inference engine. We design a class-agnostic and extensible prompting strategy to enable GPT-o3 to reason effectively using inputs from the dynamic hierarchical scene graph, current observation, action history, and task instruction. The supervisor module explicitly predicts three components for the action module: the target object, the high-level action, and a language goal description. 
The dynamic scene graph provides a comprehensive and up-to-date representation of the environment. Meanwhile, the action history, and prior language goals, supports temporal coherence, ensuring that decisions remain contextually grounded over time.

\begin{figure}[t]
\centering
\includegraphics[width=0.45\textwidth]{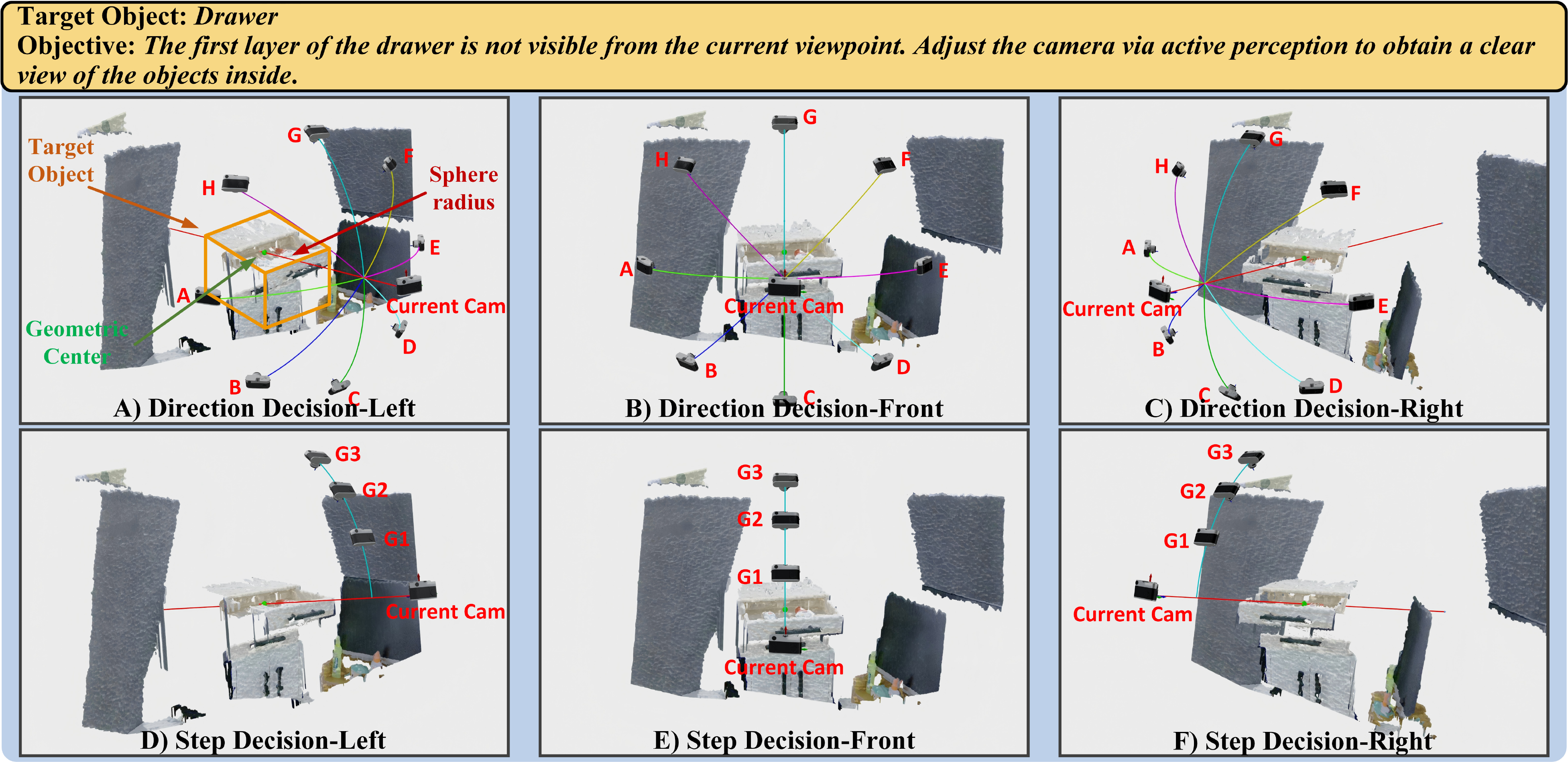} 
\caption{Illustration of active perception. Top: camera movement direction selection. Bottom: camera pose selection along the chosen direction.}
\label{fig:active_perception}
\end{figure}

\subsection{Action module}
Based on target object and high-level action, Action module invokes the corresponding sub-module to generate low-level, executable actions for the robot. In this work, we define three high-level action types: \textbf{active perception}, \textbf{interactive perception}, and \textbf{manipulation}.



Designing an \textbf{active perception} strategy guided by natural language objectives in cluttered environments is inherently challenging, as it requires interpreting semantic task goals and reasoning over the 3D scene geometry. To address this, we propose a novel visual prompting scheme that effectively leverages the reasoning capabilities of GPT-o3 to conduct task-aware scene-grounded active perception.

As illustrated in Figure \ref{fig:active_perception}, given the target object, we retrieve its 3D point cloud from memory and compute its geometric center. Using the object’s size, the camera’s field of view (FoV), and its distance from the object, we define a virtual active perception sphere centered at the object in current scene point cloud. We project the current camera pose onto the sphere and sample $N$ candidate directions along its surface—each representing a possible direction for camera movement. We then render three canonical 3D views of the scene point cloud (front, left, right) and provide them as visual prompts to GPT-o3. Guided by the action objective from the supervisor module, GPT-o3 selects the most informative direction. Next, we sample $M$ candidate poses along the selected direction, render the same canonical views, and prompt GPT-o3 to choose the optimal next camera pose. During this process, the camera remains oriented toward the object center to ensure consistent framing. Additionally, the VLM may suggest “look closer” if it infers that finer visual details are necessary to fulfill the task objective.


\textbf{Interactive perception} acquires information that cannot be obtained via passive observation, by allowing the robot to physically interact with the environment. We implement four action primitives to support interactive perception: 1) Open: Open drawers or cabinets to reveal their contents. 2) Close: Close drawers or cabinets. 3) Pick\&Place: Temporarily remove obstructing objects to uncover hidden targets. 4) Rotate: Reposition objects to expose unseen or rear-facing areas. \textbf{Manipulation} is implemented via AnyGrasp\cite{anygrasp} and action primitive. Once the target object has been identified and sufficiently perceived, the robot grasps and relocates it to a designated location, thereby completing the retrieval task.



\section{Experiments}
In this section, we evaluate the effectiveness of our proposed method in real-world tabletop scenarios using a single wrist-mounted RGB-D camera. The primary goal of our experiments is to answer the following questions: 1)
Can our system effectively perform single-camera object retrieval tasks in diverse unseen real-world tabletop scenarios? 2) How does each component of our system contribute to the overall task performance? 3) Can our method robustly handle human interventions during execution, and effectively complete multiple sequential retrieval instructions? Experiment videos are provided in the supplementary material.

\begin{table*}[t]
  \centering
  \begin{tabular}{lccccccc}
  \toprule
    \textbf{Task Category} & \textbf{Out-of-view} & \textbf{Inside} & \textbf{Behind} & \textbf{Under} & \textbf{Sequential} & \textbf{Semantic} & \textbf{Reposition} \\
  \midrule
  Hidden Inside & \ding{51} & \ding{51} & \ding{55} & \ding{55} & \ding{55} & \ding{55} & \ding{55} \\
  Recursive Search & \ding{51} & \ding{51} & \ding{51} & \ding{51} & \ding{55} & \ding{55} & \ding{55} \\
  Reposition to Reveal & \ding{51} & \ding{51} & \ding{55} & \ding{55} & \ding{55} & \ding{51} & \ding{51} \\
  Sequential Retrieval & \ding{51} & \ding{51} & \ding{55} & \ding{55} & \ding{51} & \ding{55} & \ding{55} \\
  Semantic Targeting & \ding{51} & \ding{51} & \ding{55} & \ding{55} & \ding{55} & \ding{51} & \ding{55} \\
  Compositional Reasoning & \ding{51} & \ding{51} & \ding{51} & \ding{55} & \ding{51} & \ding{51} & \ding{51} \\

  \bottomrule
  \end{tabular}
\caption{\textbf{Overview of Task Categories and Their Feature Attributes}}
\label{tab:categorized_tasks}
\end{table*}

\begin{figure}[t]
\centering
\includegraphics[width=0.45\textwidth]{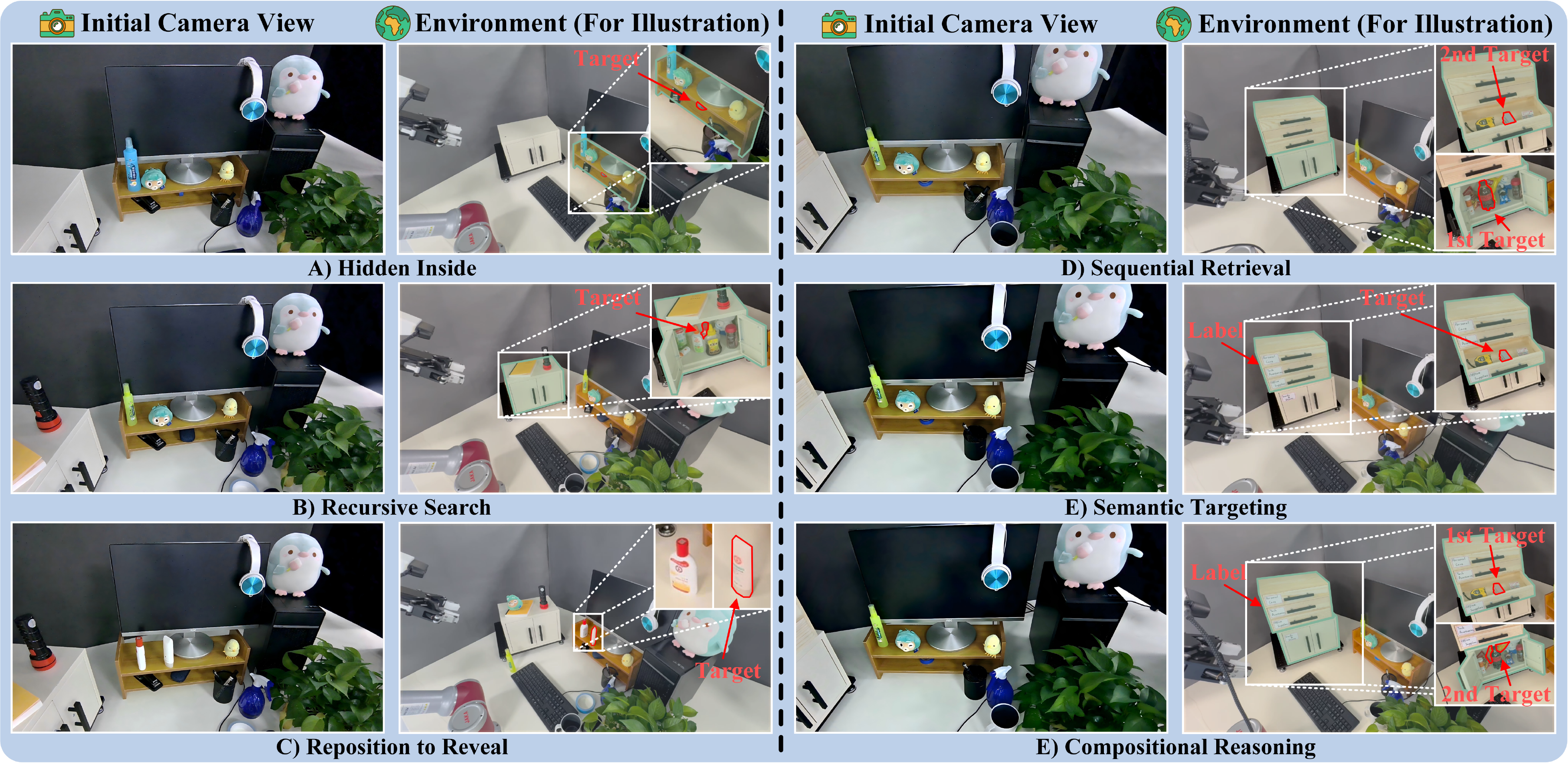} 
\caption{\textbf{Example for each task category.} Left: RGB-D observation at the beginning of the task from the \textbf{only} camera mounted on the robot wrist. Right: A third-person overview of the task environment, which is \textbf{not accessible} to the system and shown solely for illustration purposes.}
\label{fig:task_cat_example}
\end{figure}

\subsection{Experiment Setup}
To evaluate the effectiveness of our proposed method, we design six diverse task categories with varying attributes. Illustrations are shown in Figure \ref{fig:task_cat_example} and attributes are detailed in Table \ref{tab:categorized_tasks}. Each task category has three variations in task instruction, object number, type and layout.
\begin{itemize}
    \item \textbf{Hidden Inside}: Target objects may be out-of-view or hidden inside containers, need the collaboration of active perception and interactive perception to fulfill the task.
    \item \textbf{Recursive Search}: Besides out-of-view and inside, object relations such as “behind” and “under” can create unexplored area, i.e. target object can be occluded by irrelevant objects located above or in front of it. 
    \item \textbf{Reposition to Reveal}: Certain object parts may contain discriminative features that indicate their fine-grained category or function, which may be occluded or not visible from the current viewpoint. Repositioning the object enables the robot to reveal such features, especially when the scene contains multiple visually similar objects. 
    \item \textbf{Sequential Retrieval}: 
    Users may issue follow-up instructions after a retrieval task completes. This setting evaluates whether the system can reuse previous memory to avoid redundant exploration and improve efficiency.
    \item \textbf{Semantic Targeting}: In the real-world scenario, people tend to put semantic similar objects together and label them. This setting evaluates whether the system can effectively reason over semantic clues in the scene to guide more targeted exploration. 
    \item \textbf{Compositional Reasoning}: The most complicated task that combines challenges from the above categories.
\end{itemize}
All experiments are conducted in real-world tabletop environments. We use a JAKA Zu 7 robotic arm with a Gemini 336 RGB-D camera mounted on the wrist. The robot uses a DH-Robotics AG-95 parallel-jaw gripper for manipulation.

We compare our method with three baselines. \textbf{RoboExp}\cite{RoboExp}: A foundation-model-driven robotic framework that interacts with scene to explore and build scene graph, which operates with fixed camera view.  \textbf{AP-VLM}\cite{huris}: A framework that overlays a virtual 3D grid on the tabletop and guides wrist-mounted camera to focus on different grid points to locate object. \textbf{GPT-o3}: We directly feed the current RGB observation, robot pose and similar text prompt as our supervisor module into GPT-o3. For active perception, it directly generates next camera pose in text. For interactive perception and manipulation, we use manual actions as ground-truth actions. In contrast, all actions in our system are generated and executed automatically on the physical robot.

We evaluate performance using the following metrics: \textbf{Success Rate}: The success rate across all variants for each task category. An rollout is considered successful if the target object specified in the instruction is retrieved and placed at the designated location.
\textbf{Object Discovery Rate (ODR)}: The percentage of discovered objects out of all objects in the environment.
\textbf{Graph Editing Distance (GED)}: For comparison with RoboExp, we compute the distance between the ground-truth scene graph obtained from observation history and the scene graph generated by each method. The cost of adding, deleting, or moving one edge or node is 1.

\subsection{Comparison with baselines}

\begin{table*}[t]
\begin{center}
\begin{tabular}{l ccc ccc ccc}
\toprule
 & \multicolumn{3}{c}{Hidden Inside} &  \multicolumn{3}{c}{Recursive Search} & \multicolumn{3}{c}{Reposition to Reveal} \\

\cmidrule(lr){2-4} \cmidrule(lr){5-7} \cmidrule(lr){8-10}

Method 
 & Success Rate & ODR & GED & Success Rate & ODR & GED & Success Rate & ODR & GED \\

\midrule

RoboExp & 20\% & 39\% & 3.2 & 0\% & 30\% & 3.5 & 0\% & 39\% & 4.0\\

AP-VLM & 10\% & 45\% & / & 0\% & 32\% & / & 0\% & 45\% & /\\

GPT-o3 & 20\% & 43\% & / & 10\% & 38\% & / & 40\% & 65\% & /\\

\textbf{(Ours)} &
\textbf{90\%} & \textbf{95\%} & \textbf{0.3} & \textbf{80\%} &
\textbf{95\%} & \textbf{1.9} & \textbf{90\%} & \textbf{80\%} & \textbf{0.1} \\

\midrule
 & \multicolumn{3}{c}{Sequential Retrieval} &  \multicolumn{3}{c}{Semantic Targeting} & \multicolumn{3}{c}{Compositional Reasoning}\\

\cmidrule(lr){2-4} \cmidrule(lr){5-7} \cmidrule(lr){8-10}

 & Success Rate & ODR & GED & Success Rate & ODR & GED & Success Rate & ODR & GED \\

\midrule

RoboExp & 0\% & 27\% & 3.2 & 0\% & 27\% & 3.3 & 0\% & 27\% & 3.5\\

AP-VLM & 0\% & 32\% & / & 0\% & 32\% & / & 0\% & 32\% & /\\

GPT-o3 & 0\% & 30\% & / & 20\% & 30\% & / & 0\% & 30\% & /\\

\textbf{(Ours)} &
\textbf{80\%} & \textbf{88\%} & \textbf{0.1} & \textbf{90\%} &
\textbf{65\%} & \textbf{0.2} & \textbf{70\%} & \textbf{72\%} & \textbf{2.5} \\

\bottomrule
\end{tabular}
\caption{\textbf{Quantitative Results on Different Tasks.} We compare the performance of our system and baselines across various tasks. Our method consistently outperform all baselines across all metrics and tasks.}
\label{tab:Quantitative Results}
\end{center}
\end{table*}

\begin{figure*}[t]
\centering
\includegraphics[width=0.99\textwidth]{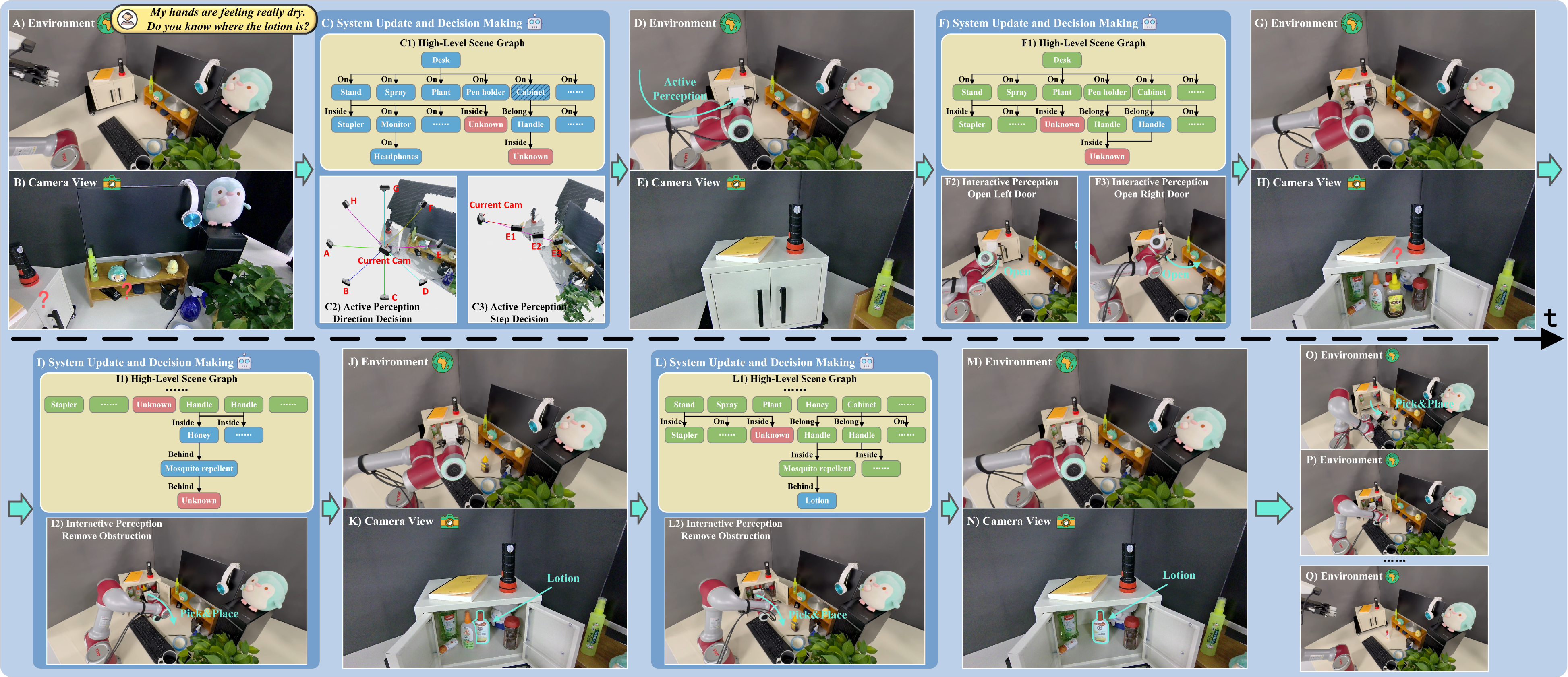} 
\caption{\textbf{Qualitative Results on the Recursive Search Task.} We present a rollout of the recursive search task, along with the high-level scene graph update and the decision-making process of our system. In the scene graph: blue nodes indicate newly added objects, green nodes represent previously existing objects, and red nodes denote unexplored regions. Nodes with diagonal stripes correspond to objects that have remained partially observed across all frames due to viewpoint limitations. For clarity and space constraints, omitted nodes are represented using ellipses $\cdots \cdots$.}
\label{fig:qualitative}
\end{figure*}

The performance of our method in comparison to all baselines is summarized in Table \ref{tab:Quantitative Results}. Our approach significantly outperforms all baselines across all task categories and evaluation metrics, demonstrating its robustness, adaptability, and effectiveness in complex, real-world scenarios under limited perception conditions. 

While RoboExp is capable of interacting with the environment, it lacks active perception capabilities. It can only perceive objects that are clearly visible from a fixed viewpoint. Given the constraint of a single wrist-mounted camera, many target objects are often located outside the current field of view or occluded, making it extremely difficult for RoboExp to complete most tasks.
AP-VLM introduces a virtual 3D grid on the tabletop, allowing for a degree of active perception by adjusting the camera's top-down view. However, this strategy lacks adaptability. It only supports constrained top-down viewpoints and cannot perform task-driven 6-DoF (6D) active perception. Furthermore, AP-VLM cannot interact with the environment, making it incapable of acquiring additional task-relevant information through physical exploration. As a result, it struggles to complete all task categories. 
Although GPT-o3 is equipped with ground-truth interactive perception and manipulation actions, its active perception generates the next camera pose based on the RGB image and robot pose. This approach fails to incorporate a grounded understanding of spatial scale within the scene. Without an anchored knowledge of the scene, the VLM often hallucinates plausible—but incorrect—viewpoints that are misaligned with the physical layout. This generative rather than discrimitive approach 
often results in meaningless or invalid camera poses. In addition, GPT-o3 lacks a memory module, preventing it from retaining scene history or maintaining task consistency over time. 

Figure~\ref{fig:qualitative} illustrates a rollout example of the \textbf{Recursive Search} task, highlighting our system’s scene graph updates and decision-making process.
In this scenario, the user requests retrieval of a \textbf{\textit{lotion}}. After the initial observation, the system hypothesizes that the lotion might be located inside either the \textbf{\textit{cabinet}} or the \textbf{\textit{stand}}, and decides to explore the cabinet first. In the initial camera viewpoint, the cabinet is only partially visible. The system then autonomously adjusts its viewpoint through active perception, achieving a clear view of the cabinet. Next, it performs interactive perception by opening the cabinet to inspect its interior. However, occlusions among objects inside create new unexplored areas. By recursively invoking interactive perception, the system removes obstructions, locates the lotion, and places it at the designated location. 
This example highlights our method’s ability to reason over spatial relationships, dynamically update its internal representation of the scene, and coordinate active and interactive perception to accomplish complex object retrieval tasks under limited perceptual conditions.

\subsection{Ablation Study}

\begin{figure}[t]
\centering
\includegraphics[width=0.48\textwidth]{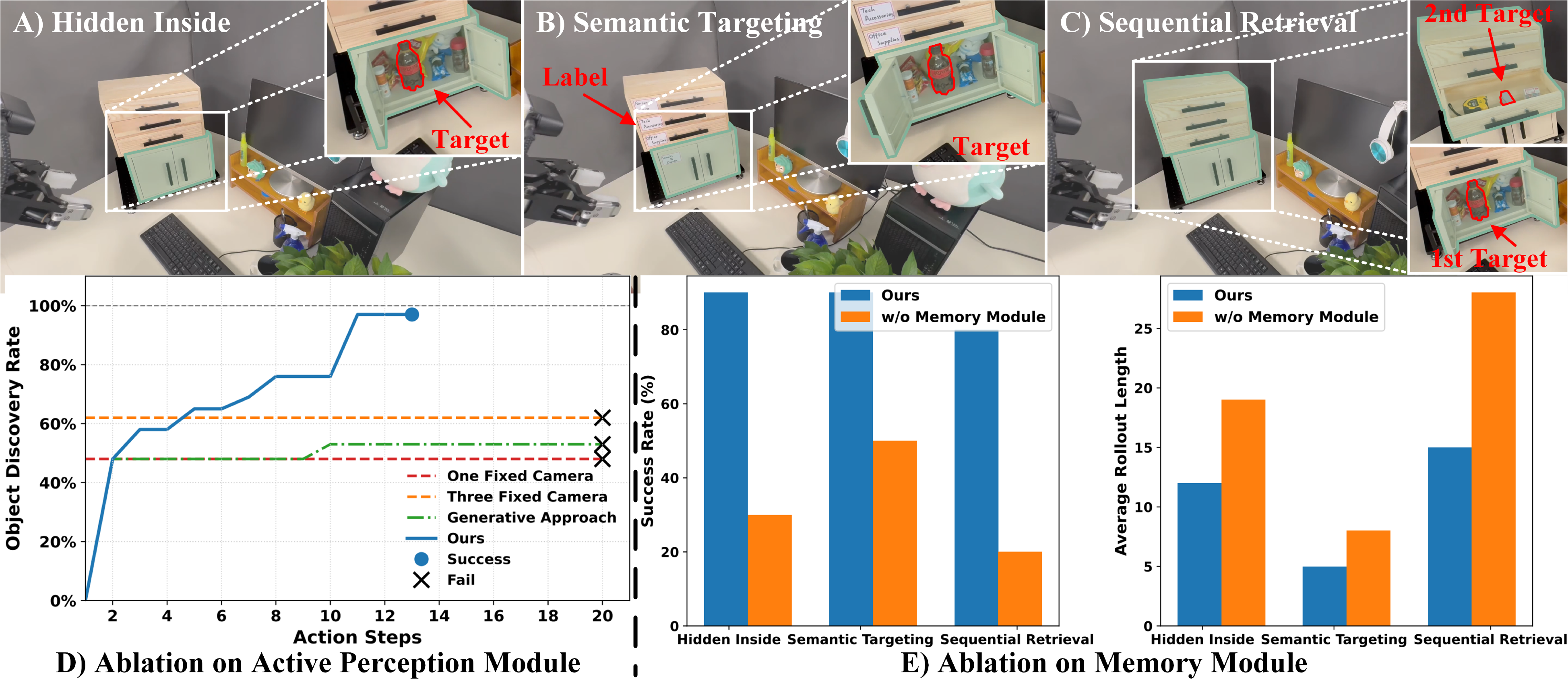} 
\caption{(A-C) Task variants used in the ablation study. (D) Ablation results for our active perception module compared with baselines. (E) Ablation results for the memory module.}
\label{fig:ablation}
\end{figure}

In this section, we conduct an ablation study to validate the effectiveness of our proposed active perception and memory modules.
Experiments are performed on specific variants of three task categories: Hidden Inside, Semantic Targeting, and Sequential Retrieval, as illustrated in Figure \ref{fig:ablation} A, B, C. The environments for Hidden Inside and Sequential Retrieval are identical, except that Sequential Retrieval involves a sequence of user-issued instructions. For Semantic Targeting, the environment differs only by including semantic labels for each drawer layer or cabinet. For example, a cabinet containing items like cola, Oreos, and chips is labeled “Snacks \& Drinks”.

We first assess the effectiveness of our proposed active perception module on the Hidden Inside task. 
Figure~\ref{fig:ablation} D shows the object discovery rate over time for our full method compared to three ablated baselines: (1) Using a single fixed RGB-D camera. (2) Using three fixed RGB-D cameras positioned at different viewpoints. (3) Using a generative strategy similar to GPT-o3 baseline, predicting the next camera pose purely based on RGB image and robot pose. 
As illustrated, our method continuously discovers new objects throughout task execution, demonstrating the effectiveness of our proposed task-aware scene-grounded active perception. In contrast, all baselines struggle to go beyond the initial visible objects set due to the lack of exploratory ability.

Next, we evaluate the importance of our memory module and semantic understanding ability across three ablation tasks, as shown in Figure \ref{fig:ablation} E. Removing the memory module leads to a substantial drop in success rate, indicating that persistent memory of scene and action history are critical for maintaining task-relevant context, especially under partial observability.
Additionally, our method achieves significantly shorter average rollout lengths in the Semantic Targeting task compared to the Hidden Inside task. This indicates that the system can effectively leverage semantic labels in the environment to guide more targeted exploration. In Sequential Retrieval, our method also outperforms the method without memory module in terms of rollout length. This shows that the memory module enables the agent to reuse previous scene knowledge, improving efficiency in sequential retrieval tasks. Lastly, we also verified the robustness of our method under human interventions. More details can be found in the supplementary video.

\section{Conclusion}
In this work, we introduced RoboRetriever, a novel robotic framework that effectively integrates active and interactive perception for robust object retrieval using only a single wrist-mounted camera. By leveraging a dynamic hierarchical scene graph, a task-aware scene-grounded prompting scheme, and a reasoning VLM, our method enables robots to autonomously reason about unexplored or occluded regions, iteratively update their internal memory, and execute task-guided perception and manipulation actions based on natural language instructions. Extensive real-world experiments across diverse scenarios demonstrated that RoboRetriever significantly outperforms existing methods.
Future work will explore the incorporation of multi-arm coordination 
to further expand the system's capability.

\bibliography{aaai2026}

\end{document}